\documentclass[letter, 12pt]{article}
\usepackage{arxiv}

\usepackage[utf8]{inputenc} % allow utf-8 input
\usepackage[T1]{fontenc}    % use 8-bit T1 fonts
\usepackage{hyperref}       % hyperlinks
\usepackage{url}            % simple URL typesetting
\usepackage{booktabs}       % professional-quality tables
\usepackage{amsfonts}       % blackboard math symbols
\usepackage{nicefrac}       % compact symbols for 1/2, etc.
\usepackage{microtype}      % microtypography
\usepackage{lipsum}		% Can be removed after putting your text content
\usepackage{graphicx}
\usepackage{natbib}
\usepackage{doi}
\usepackage{caption}
\usepackage{multirow}
\usepackage{makecell}

\usepackage{mathptmx}
\usepackage[T1]{fontenc}
\title{Transformer-Based Named Entity Recognition for French Using Adversarial Adaptation to Similar Domain Corpora}

\author{{Arjun Choudhry$^*$, Pankaj Gupta$^*$, Inder Khatri, Aaryan Gupta}\\
	Biometric Research Laboratory\\
	Delhi Technological University, New Delhi, India\\
	\texttt{ \{choudhry.arjun, pankajgupta.dtu, inderkhatri999, aryan227227\}@gmail.com}\\
	$^*$These authors contributed equally.\\
    \AND {Maxime Nicol}\\
	 Université du Québec à Montréal\\
	 Montréal, QC, Canada\\
    \texttt{nicol.maxime@courrier.uqam.ca} \\
	\And
	\href{https://orcid.org/0000-0001-8196-2153}{\includegraphics[scale=0.06]{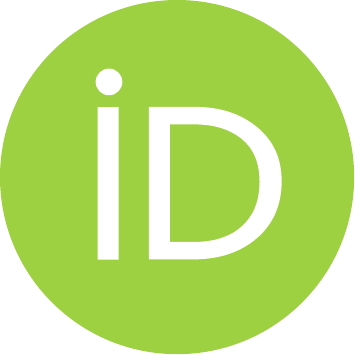}\hspace{1mm}Marie-Jean Meurs} \\
	Université du Québec à Montréal\\
	Montreal, QC, Canada\\
	\texttt{meurs.marie-jean@uqam.ca} \\
    \And{Dinesh Kumar Vishwakarma}\\
    Biometric Research Laboratory\\
	Delhi Technological University\\
	New Delhi, India\\
	\texttt{dinesh@dtu.ac.in}\\
}

\date{}

\renewcommand{\shorttitle}{Transformer-Based NER for French\\Using Adversarial Adaptation to Similar Domain Corpora}

\hypersetup{
pdftitle={Transformer-Based Named Entity Recognition for French Using Adversarial Adaptation to Similar Domain Corpora},
pdfsubject={cs.CL},
pdfauthor={Arjun Choudhry, Pankaj Gupta, Inder Khatri, Aaryan Gupta, Maxime Nicol, Marie-Jean Meurs, Dinesh Kumar Vishwakarma},
pdfkeywords={Named Entity Recognition, French Corpora, Adversarial Adaptation, Information Retrieval, Natural Language Processing},
}

\begin{document}
\maketitle

\begin{abstract}
Named Entity Recognition (NER) involves the identification and classification of named entities in unstructured text into predefined classes. NER in languages with limited resources, like French, is still an open problem due to the lack of large, robust, labelled datasets. In this paper, we propose a transformer-based NER approach for French using adversarial adaptation to similar domain or general corpora for improved feature extraction and better generalization. We evaluate our approach on three labelled datasets and show that our adaptation framework outperforms the corresponding non-adaptive models for various combinations of transformer models, source datasets and target corpora.
\end{abstract}

% keywords can be removed

\section{Introduction}

Named Entity Recognition (NER) is an information extraction task where specific entities are extracted from unstructured text and labelled into predefined classes. While NER models for high-resource languages like English have seen notable performance gains due to improvements in model architectures and availability of large datasets, limited-resource languages like French still face a dearth of openly available, large, labelled datasets. Recent research works use adversarial adaptation frameworks for adapting NER models from high-resource domains to low-resource domains. These approaches have been used for high-resource languages, where robust language models are available. We utilize adversarial adaptation to enable models to learn better, generalized features by adapting them to large, unlabelled corpora for better performance on source test set. 

We propose a Transformer-based NER approach for French using adversarial adaptation to counter the lack of large, labelled NER datasets in French. We train transformer-based NER models on labelled source datasets and use larger corpora from similar or mixed domains as target sets for improved feature learning. Our proposed approach helps outsource wider domain and general feature knowledge from easily-available large, unlabelled corpora. While we limit our evaluation to French datasets and corpora, our approach can be applied to other languages too. 

\section{Proposed Methodology}

\begin{figure*}[t!]
     \centering
     \includegraphics[width = \textwidth]{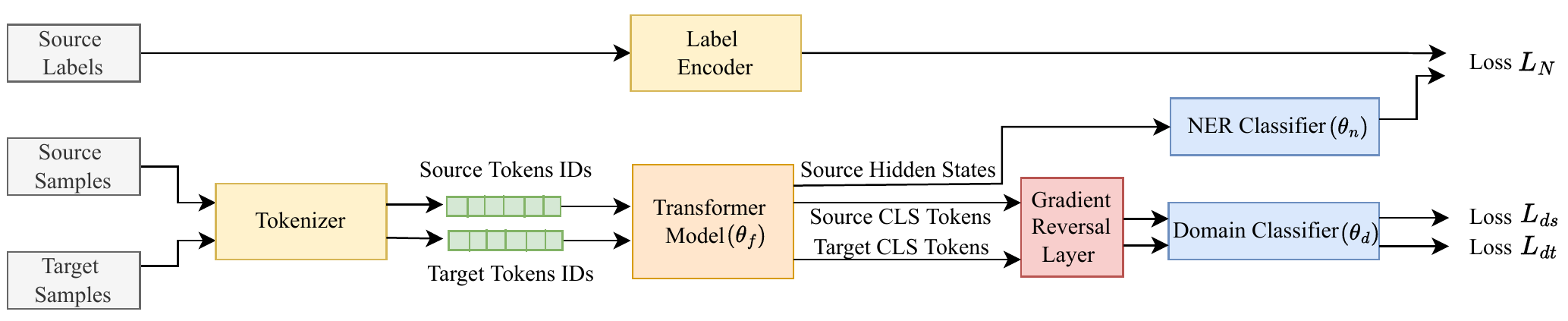}
     \hfill
   \caption{Graphical representation of our adversarial adaptation framework for training NER models on source and target sets.}
   \label{Flowchart} 
\end{figure*}

\subsection{Datasets and Preprocessing}
We use WikiNER French~\cite{wikiner}, WikiNeural French~\cite{wikineural}, and Europeana French~\cite{Europeana} datasets as the labelled source datasets in our work. Europeana is extracted from historic European newspapers using Optical Character Recognition (OCR), and contains OCR errors, leading to a noisy dataset. For unlabelled target corpora, we use the WikiNER and WikiNeural datasets without their labels, and the Leipzig Mixed French (Mixed-Fr) corpus\footnotemark[1]. 
These enable us to evaluate the impact of adapting models to similar domain, as well as generalized corpora. During preprocessing, we convert all NER tags to Inside-Outside-Beginning (IOB) format~\cite{IOB}.

\footnotetext[1]{https://wortschatz.uni-leipzig.de/en/download/French}

\subsection{Adversarial Adaptation to Similar Domain Corpus}
Adversarial adaptation helps select domain-invariant features transferable between source and target datasets \cite{Adversarial_adapt}. Based on this premise, we propose that adversarially adapting NER models to large, unlabelled corpora from similar domain as the source helps enable the model to extract more generalizable features. This reduces overfitting on the intricate training set-specific features. We also test the same for the case where target dataset is a mixed-domain, large corpus. We test our approach for three conditions: source and target datasets are from the same domain; source and target datasets are from relatively different domains; and target dataset is a mixed-domain, large-scale, general corpus. Figure~\ref{Flowchart} illustrates our proposed framework. The domain classifier acts as a discriminator. NER classifier loss, adversarial loss, and total loss are defined as: 

\begin{equation}
%  \small
%  \scriptstyle 
 L_{NER} \ = \ \min\limits_{\theta_{f},\theta_{n}} \sum_{i=1}^{n_{s}} L_{n}^i
\end{equation}
 
\begin{equation}
%  \small
%  \scriptstyle 
 L_{adv} \ = \ \min\limits_{\theta_d} (\max\limits_{\theta_f}( \sum_{i=1}^{n_{s}} L_{ds}^i \ + \ \sum_{j=1}^{n_{t}} L_{dt}^j))
\end{equation}

\begin{equation}
%  \small
%  \scriptstyle 
 L_{Total} \ = \ L_{NER} \ + \ \alpha (L_{adv})
\end{equation}

where $n_s$ and $n_t$ are number of samples in source and target sets, $\theta_d$, $\theta_n$ and $\theta_f$ are number of parameters for domain classifier, NER classifier and transformer model, $L_{d_s}$ and $L_{d_t}$ represent the Negative log likelihood loss for source and target respectively, and $\alpha$ is ratio between $L_{NER}$ and $L_{adv}$. We found $\alpha = 2$ to provide the best experimental results.

\subsection{Language Models for NER}
% \vspace{-.24cm}
Recent NER research has incorporated large language models due to their contextual knowledge learnt during pretraining~\cite{xlnet-bilstm-crf,chinese-bert,transformer-NER}. We use three French language models for evaluating our proposed approach: CamemBERT-base \cite{camembert}, CamemBERT-Wiki-4GB (a variant of CamemBERT pretrained on only 4GB of Wikipedia corpus), and FlauBERT-base \cite{flaubert}. Comparing CamemBERT-base and CamemBERT-Wiki-4GB helps us analyse if we can replace large language models with smaller ones adapted to unlabelled corpora during fine-tuning on a downstream task.

\section{Experimental Results and Discussion}
We evaluated our approach on various combinations of language models, source and target datasets. Each model was evaluated on the test set of source dataset. Table \ref{DANER-Fr_results} illustrates our results. Some findings observed are described hereafter.

Adversarial adaptation models outperform their non-adaptive counterparts: 
We observed that the adaptation models consistently outperformed their non-adaptive counterparts across almost all combinations of datasets and language models on precision, recall and F1-score. 

Adversarial adaptation can help alleviate performance loss on using smaller models: 
Fine-tuning CamemBERT-Wiki-4GB using our adversarial approach helped achieve similar performance to non-adapted CamemBERT-base for certain datasets. CamemBERT-Wiki-4GB adapted to WikiNeural corpus even outperformed unadapted CamemBERT-base for WikiNER dataset. Thus, adversarial adaptation during fine-tuning could act as a substitute for using larger language models.

Adapting models to same domain target corpora leads to slightly better performance than adapting to a mixed corpus: 
We observed that models adapted to corpora from same domain as source dataset (like for WikiNER and WikiNeural as source and target datasets, or vice versa) showed equal or slightly better performance than models adapted to general domain.

Adapting models to mixed-domain target corpus leads to better performance than adapting to a corpus from a different domain: 
We observed that models adapted to mixed-domain corpora (Europeana to Mixed-Fr) showed noticeably better performance than models adapted to corpora from different domains (Europeana to WikiNER).

\begin{table}[t!]
    \centering
    % \large
    % \scale{.2}
    \resizebox{.7\columnwidth}{!}{
    \begin{tabular}{c|c|c|ccc}
    % \hline
    % \hline
        Model & Source & Target & Precision & Recall & F1 \\\hline
        \multirow{9}{*}{\makecell{CamemBERT-\\Wiki-4GB}}& \multirow{3}{*}{\makecell{WikiNER}}& &  0.911& 0.925& 0.918\\
        & & WikiNeural&  \textbf{0.966}& \textbf{0.963}& \textbf{0.969}\\
        & & Mixed-Fr& 0.956& 0.962& 0.959\\
    \cline{2-6}
        & \multirow{3}{*}{\makecell{WikiNeural}}& & 0.859& 0.872& 0.866\\
        & & WikiNER & \textbf{0.872}& \textbf{0.891}& \textbf{0.881}\\
        & & Mixed-Fr& 0.870 & 0.879 & 0.875\\
    \cline{2-6}
        & \multirow{3}{*}{\makecell{Europeana}}&  & 0.728 & 0.642 & 0.682\\
        & & WikiNER& 0.738& \textbf{0.691}& \textbf{0.714}\\
        & & Mixed-Fr& \textbf{0.774} & 0.640 & 0.701\\
    \hline
        \multirow{9}{*}{\makecell{CamemBERT-\\base}}& \multirow{3}{*}{\makecell{WikiNER}}& & 0.960& 0.968& 0.964\\
        & & WikiNeural& \textbf{0.973}& 0.976& \textbf{0.975}\\
        & & Mixed-Fr& 0.972& \textbf{0.978}& 0.974\\
    \cline{2-6}
        & \multirow{3}{*}{\makecell{WikiNeural}}& & 0.943& 0.950& 0.946\\
        & & WikiNER& 0.943& \textbf{0.953} & \textbf{0.948}\\
        & & Mixed-Fr & \textbf{0.946} & 0.950 & \textbf{0.948}\\
    \cline{2-6}
        & \multirow{3}{*}{\makecell{Europeana}}& & 0.927 & 0.933 & 0.930\\
        & & WikiNER& 0.911& 0.927& 0.920\\
        & & Mixed-Fr& \textbf{0.942} & \textbf{0.943} & \textbf{0.943}\\
    \hline
        \multirow{9}{*}{\makecell{FlauBERT-\\base}}& \multirow{3}{*}{\makecell{WikiNER}}& & 0.963& 0.964& 0.963\\
        & & WikiNeural& 0.964& 0.968& 0.966\\
        & & Mixed-Fr& \textbf{0.974}& \textbf{0.972}& \textbf{0.973}\\
    \cline{2-6}
        & \multirow{3}{*}{\makecell{WikiNeural}}& & 0.934& 0.946& 0.940\\
        & & WikiNER& 0.935& \textbf{0.950} & \textbf{0.942}\\
        & & Mixed-Fr& \textbf{0.941} & 0.943 & \textbf{0.942}\\
    \cline{2-6}
        & \multirow{3}{*}{\makecell{Europeana}}& & 0.835 & 0.863 & 0.849\\
        & & WikiNER& 0.855& \textbf{0.865}& 0.860\\
        & & Mixed-Fr& \textbf{0.882} & 0.854 & \textbf{0.867}\\
    % \hline
    % \hline
    \end{tabular}
    }
    
    \caption{Performance evaluation of our proposed approaches for various combinations of models, source and target sets.}
    \label{DANER-Fr_results}
\end{table}

\textbf{Reproducibility.~}
The \href{https://github.com/Arjun7m/AA_NER_Fr}{source code of the proposed systems} is licensed under the GNU~GPLv3. 
The datasets are publicly available.

\textbf{Acknowledgments.~}
This research was enabled by support provided by \href{https://www.calculquebec.ca}{Calcul Québec}, \href{https://www.alliancecan.ca}{The Alliance} and \href{https://www.mitacs.ca}{MITACS}.

\bibliographystyle{unsrtnat}
\bibliography{aaai2023_SA_arXiv_version}

\end{document}